\documentclass{article}
\usepackage{ijcai25}

\usepackage{times}
\usepackage{soul}
\usepackage{url}
\usepackage[hidelinks]{hyperref}
\usepackage[utf8]{inputenc}
\usepackage[small]{caption}
\usepackage{graphicx}
\usepackage{amsmath}
\usepackage{amsthm}
\usepackage{booktabs}
\usepackage{algorithm}
\usepackage{algorithmic}
\usepackage[switch]{lineno}

\usepackage{bm}
\usepackage{arydshln} 
\usepackage{tikz}
\usetikzlibrary{fadings}
\usetikzlibrary{mindmap,trees}
\usetikzlibrary{arrows,automata,shapes,positioning,shadows,trees}
\usetikzlibrary{shapes,snakes,shadows}
\usetikzlibrary{shapes.arrows}
\usetikzlibrary{calc,shapes, positioning}
\usetikzlibrary{decorations.text}
\usetikzlibrary{matrix,chains,positioning,decorations.pathreplacing,arrows}
\usetikzlibrary{bayesnet}
\usepackage[edges]{forest}
\usetikzlibrary{shadows.blur}
\usetikzlibrary{shapes.geometric}

\title{One-shot Federated Learning Methods: A Practical Guide}

\author{
Xiang Liu$^{1}$\and
Zhenheng Tang$^{2}$ \and
Xia Li$^{3}$\and 
Yijun Song$^4$ \and 
Sijie Ji$^5$\and \\
Zemin Liu$^6$\and 
Bo Han$^{7,\dagger}$ \and
Linshan Jiang$^{1}$\And
Jialin Li$^{1}$ 
\affiliations
$^1$National University of Singapore,
$^2$The Hong Kong University of Science and Technology,\\
$^3$Department of Information Technology and Electrical Engineering, ETH Zürich,\\
$^4$Zhejiang University of Finance \& Economics Dongfang College, 
$^5$California Institute of Technology\\
$^6$College of Computer Science and Technology, Zhejiang University,\\
$^7$Department of Computer Science, Hong Kong Baptist University\\
$\dagger$ Corresponding Authors
\emails
\{liuxiang,lijl\}@comp.nus.edu.sg, zhtang.ml@ust.hk, xiaxiali@ethz.ch, yijunsong@zufedfc.edu.cn,\\
sijieji@caltech.edu, liu.zemin@zju.edu.cn, bhanml@comp.hkbu.edu.hk, linshan@nus.edu.sg
}
\begin{document}

\maketitle

\begin{abstract}
One-shot Federated Learning (OFL) is a distributed machine learning paradigm that constrains client-server communication to a single round, addressing privacy and communication overhead issues associated with multiple rounds of data exchange in traditional Federated Learning (FL). OFL demonstrates the practical potential for integration with future approaches that require collaborative training models, such as large language models (LLMs). However, current OFL methods face two major challenges: data heterogeneity and model heterogeneity, which result in subpar performance compared to conventional FL methods. Worse still, despite numerous studies addressing these limitations, a comprehensive summary is still lacking. To address these gaps, this paper presents a systematic analysis of the challenges faced by OFL and thoroughly reviews the current methods. We also offer an innovative categorization method and analyze the trade-offs of various techniques. Additionally, we discuss the most promising future directions and the technologies that should be integrated into the OFL field. This work aims to provide guidance and insights for future research.
\end{abstract}

\section{Introduction}
Federated Learning (FL)~\cite{fedavg1} is an established paradigm that fosters multiple clients to collaboratively engage in distributed machine learning, allowing the aggregation of local models on a server to generate a global model. In practice, current implementations of FL require multiple rounds of data exchange between clients and the server to ensure the training of a high-accuracy global model without the need to directly utilize raw data. However, as FL is widely employed in fields such as intelligent transportation, economy, manufacturing and healthcare~\cite{kairouz2021advances}, the traditional multi-round setup can still violate privacy-preserving principles~\cite{kairouz2021advances} and incur significant communication overhead~\cite{wang2023data,tang2024fusefl}.

To protect data privacy, researchers have proposed three main techniques to address privacy leakage in FL: Differential Privacy (DP) \cite{dwork2011differential}, Secure Multi-Party Computation (SMPC) \cite{bonawitz2017practical}, and Homomorphic Encryption (HE) \cite{cheon2017homomorphic}. DP-based methods introduce noise into the intermediate data before sharing to prevent privacy leakage, but this can reduce model accuracy. SMPC-based methods increase network communication overhead between participants, especially when scaling to a large number of clients. HE-based models are criticized for requiring substantial computational resources. While these methods address privacy-preserving issues, they do not tackle communication overhead and can introduce additional issues. 

Consequently, communication between server and clients is limited to a single round, as in one-shot Federated Learning (OFL)~\cite{guha2019one}.
OFL methods have been specifically designed to improve accuracy in one-shot scenarios. OFL not only reduces the burden of communication transmission but also potentially achieves even stronger security due to its single-round setting when integrated with existing privacy-preserving methods. Additionally, it offers further advantages, such as alleviating the requirements for transmission synchronization~\cite{alemdar2021rfclock}. Therefore, OFL methods have emerged as a promising approach to address these practical issues. Since 2019, a significant number of researchers, recognizing the potential of OFL, have begun to study this area and have already proposed various methods aimed at addressing specific challenges as well as enhancing global model performance on the server side~\cite{tang2024fusefl,shen2025bridging}.

However, none of the current papers has effectively summarized the challenges within the OFL domain, and most researchers have provided inadequate summaries and classifications of OFL-related literature. This may be due to the rapid development of OFL as a practical field, which has yet to receive a thorough overview. Additionally, because current OFL methods often employ multiple techniques simultaneously, categorization becomes muddled, leading to inconsistent and imprecise classification approaches in the literature.

Existing survey papers tend to focus on Knowledge Distillation~\cite{mora2022knowledge}, Label Leakage~\cite{liu2024label}, or integration with Foundation Models~\cite{woisetschlager2024survey} in FL, but none have concentrated on providing a comprehensive overview of the challenging and promising practical methods in OFL. As Large Language Models (LLMs) gain prominence, maintaining data privacy during large-scale model training and reducing communication overhead through collaborative training among multiple clients becomes increasingly important~\cite{tang2024fusionllmdecentralizedllmtraining,openfedllm}. Especially with the development of cloud-edge collaborative frameworks~\cite{tang2024fusionllmdecentralizedllmtraining}, OFL may demonstrate even broader applicability alongside LLMs.

Thus, we present the first survey paper specifically focused on the OFL domain. Our main contributions are not limited to a standard survey; they are summarized as follows:

\begin{itemize}
    \item We propose novel taxonomies addressing the challenges faced by OFL and current methods, aiding researchers in better understanding the issues and existing approaches.

    \item We provide a thorough and detailed overview of existing OFL methods, including their primary designs, underlying principles, and advantages and potential drawbacks in addressing specific challenges. This enables researchers to gain a deeper understanding of approaches.

    \item Beyond summarizing existing methods, we discuss our findings, highlight promising future directions, and identify relevant areas, assisting researchers in integrating OFL with practical applications.
\end{itemize}

The paper is organized as follows. Section~\ref{sec:challenge} introduces the fundamental challenges of one-shot Federated Learning. Section~\ref{sec:method} provides a novel taxonomy and discusses in detail the technical aspects of each category within the context of OFL. Section~\ref{sec:future} offers a discussion that summarizes our findings and outlines future directions for the development of OFL methods. Finally, Section~\ref{sec:conclusion} concludes our paper.

\section{Challenges in One-shot Federated Learning}
\label{sec:challenge}
\subsection{Problem Formulation}
In federated learning, clients collaboratively train local models using their private local datasets to produce a global model for the server within a distributed fashion. The objective is to minimize the optimization problem:

\begin{equation}
        \label{obj}
        \min\limits_{\bm{w}} F(\bm{w}) :=  \alpha_i \sum_{i \in [n]}  F_i(\bm{w_i})
\end{equation}

where $n$ denotes the number of clients, $\bm{w}_i$ denotes the parameters of local client$_i$, $\bm{w}$ denotes the parameters of the global model, $\alpha_i$ denotes the proportion of client$_i$'s private dataset to the entire dataset, $F$ is the loss function.

In the traditional FL setting, clients upload their parameters $\bm{w}_i$ multiple times to get the final $\bm{w}$. However, in OFL, they only upload their parameters once.

\subsection{Proposed Taxonomy for Challenges}

Besides the privacy issues, OFL methods must improve the global model's test accuracy while addressing two main challenges: (1) \textbf{Data Heterogeneity}, where the private datasets on each client are non-iid (independent and identically distributed); and (2) \textbf{Model Heterogeneity}, as the local models participating in FL often differ from one another.

\subsubsection{Data Heterogeneity}
In practice, data heterogeneity arises from the non-IID characteristics of the private datasets among clients, which are primarily due to three types of skew~\cite{li2022federated}: quantity skew, feature skew and label skew.

For quantity skew, each client has the same data distribution but with different amounts of data. For feature skew, the classes within different clients' private datasets have different statistics. For label skew, different clients have varying proportions of data for the same class. 

In traditional FL, numerous methods have been proposed to handle data heterogeneity, such as FedAvg~\cite{fedavg1} (averaging parameters from clients), SCAFFOLD~\cite{scaffold} and MOON~\cite{li2021model} (measuring the similarity and disparity of parameters between parties), FedProx~\cite{fedprox} (adding an L2 regularization term for optimization) and VHL~\cite{VHL} (calibrating features via virtual data). These methods work well in non-one-shot scenarios by iteratively updating and integrating the features of client models with the server round by round~\cite{li2022federated}.

However, in a one-shot setting with non-IID data, the differences in model data across clients can be substantial, posing new challenges that traditional methods struggle to address efficiently in a single pass. This challenge is essentially an out-of-distribution (OOD) detection~\cite{yang2024generalized} problem. Since models are uploaded only once, if the FL method fails to capture all the hidden dataset statistics in the client parameters effectively, parts of the data samples may be OOD in the global model and remain unrecognized~\cite{guha2019one}. In multi-round settings, global models can be finetuned through iterative updates, but OFL methods require more robust learning capabilities to overcome the data heterogeneity challenge without sacrificing accuracy.

\subsubsection{Model Heterogeneity}
In practice, having all clients and the server share the same model as the global model often fails to meet the diverse needs of all clients, especially since FL is commonly integrated with different model architectures within cloud-edge collaborative frameworks or deployed on different edge devices, or due to privacy considerations~\cite{zhang2024does}. Consequently, personalized FL~\cite{10.5555/3294996.3295196} was proposed to address these concerns, and most of this research is based on the assumption of model homogeneity.

Model heterogeneity arises from differences in the communication capabilities~\cite{Shah2021ModelCF}, computing resources~\cite{shin2024effective}, and model architectures of local clients~\cite{shen2025bridging}. Even when clients share the same model, disparities in computing resources can lead to different clients using different sub-models from the global model~\cite{diao2021heterofl,shen2025bridging} for training or updating only specific layers' parameters to meet varying requirements~\cite{li2021fedbn}, resulting in model heterogeneity. Additionally, clients may be unwilling to disclose details of their model designs, further complicating the issue. 

Therefore, in OFL tasks, researchers must consider not only data heterogeneity but also the persistent challenge of model heterogeneity. Current approaches for model heterogeneity of OFL primarily concentrate on addressing the challenge posed by differing model architectures among clients.

\definecolor{connect-line}{RGB}{0,0,0}
\definecolor{middle-color}{RGB}{255,255,255}
\definecolor{leaf-color}{RGB}{255,255,255}
\definecolor{line-color}{RGB}{25,25,112}

\definecolor{black}{RGB}{0,0,0}


\definecolor{pure}{RGB}{112,25,25}
\definecolor{node}{RGB}{25,25,112}
\definecolor{graph}{RGB}{25,112,25}
\definecolor{graph1}{RGB}{237,145,33}
\definecolor{graph2}{RGB}{160,102,211}

\tikzstyle{pure-leaf}=[draw=pure,
    rounded corners,minimum height=1em,
    fill=leaf-color!40,text opacity=1, align=center,
    fill opacity=.5,  text=black,align=left,font=\scriptsize,
    inner xsep=3pt,
    inner ysep=1pt,
]
\tikzstyle{pure-middle}=[draw=pure,
    rounded corners,minimum height=1em,
    fill=middle-color!40,text opacity=1, align=center,
    fill opacity=.5,  text=black,align=left,font=\scriptsize,
    inner xsep=3pt,
    inner ysep=1pt,
]

\tikzstyle{node-leaf}=[draw=node,
    rounded corners,minimum height=1em,
    fill=leaf-color!40,text opacity=1, align=center,
    fill opacity=.5,  text=black,align=left,font=\scriptsize,
    inner xsep=3pt,
    inner ysep=1pt,
]
\tikzstyle{node-middle}=[draw=node,
    rounded corners,minimum height=1em,
    fill=middle-color!40,text opacity=1, align=center,
    fill opacity=.5,  text=black,align=left,font=\scriptsize,
    inner xsep=3pt,
    inner ysep=1pt,
]

\tikzstyle{graph-leaf}=[draw=graph,
    rounded corners,minimum height=1em,
    fill=leaf-color!40,text opacity=1, align=center,
    fill opacity=.5,  text=black,align=left,font=\scriptsize,
    inner xsep=3pt,
    inner ysep=1pt,
]
\tikzstyle{graph-middle1}=[draw=graph1,
    rounded corners,minimum height=1em,
    fill=middle-color!40,text opacity=1, align=center,
    fill opacity=.5,  text=black,align=left,font=\scriptsize,
    inner xsep=3pt,
    inner ysep=1pt,
]
\tikzstyle{graph-middle2}=[draw=graph2,
    rounded corners,minimum height=1em,
    fill=middle-color!40,text opacity=1, align=center,
    fill opacity=.5,  text=black,align=left,font=\scriptsize,
    inner xsep=3pt,
    inner ysep=1pt,
]

\tikzstyle{graph-middle}=[draw=graph,
    rounded corners,minimum height=1em,
    fill=middle-color!40,text opacity=1, align=center,
    fill opacity=.5,  text=black,align=left,font=\scriptsize,
    inner xsep=3pt,
    inner ysep=1pt,
]

\tikzstyle{leaf}=[draw=line-color,
    rounded corners,minimum height=1em,
    fill=leaf-color!40,text opacity=1, align=center,
    fill opacity=.5,  text=black,align=left,font=\scriptsize,
    inner xsep=3pt,
    inner ysep=1pt,
    ]
\tikzstyle{middle}=[draw=line-color,
    rounded corners,minimum height=1em,
    fill=middle-color!40,text opacity=1, align=center,
    fill opacity=.5,  text=black,align=left,font=\scriptsize,
    inner xsep=3pt,
    inner ysep=1pt,
    ]
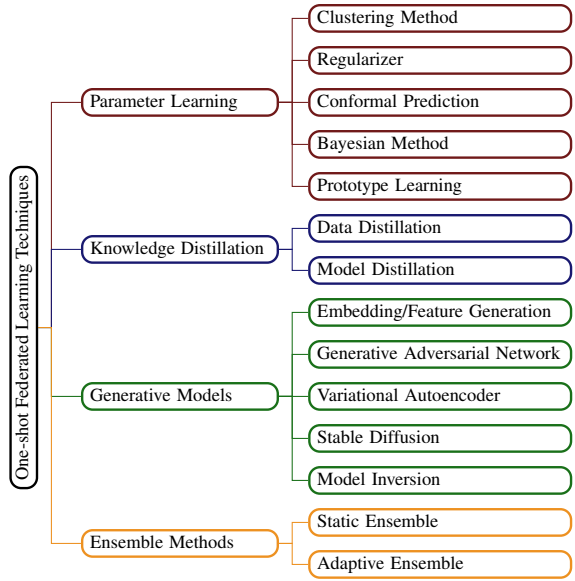
\begin{figure}[!ht]
\centering

\begin{forest}
  for tree={
    forked edges,
    grow=east,
    reversed=true,
    anchor=base west,
    parent anchor=east,
    child anchor=west,
    base=middle,
    font=\scriptsize,
    rectangle,
    line width=0.9pt,
    draw=connect-line,
    rounded corners,align=left,
    minimum width=2em,
    s sep=5pt,
    inner xsep=3pt,
    inner ysep=1pt,
  },
  where level=1{text width=6em}{},
  where level=2{text width=6em,font=\scriptsize}{},
  where level=3{font=\scriptsize}{},
  where level=4{font=\scriptsize}{},
  where level=5{font=\scriptsize}{},
  [One-shot Federated Learning Techniques, black, rotate=90,anchor=north,edge=pure
    [Parameter Learning, pure-middle, edge=pure, text width=6.8em
        [Clustering Method, pure-middle, text width=9.3em, edge=pure]
        [Regularizer, pure-middle, text width=9.3em, edge=pure]
        [Conformal Prediction, pure-middle, text width=9.3em, edge=pure
        ]
        [Bayesian Method, pure-middle, text width=9.3em, edge=pure]
        [Prototype Learning, pure-middle, text width=9.3em, edge=pure]
    ]
    [Knowledge Distillation, node-middle, edge=node, text width=6.8em
        [Data Distillation, node-middle, edge=node,text width=9.3em
        ]
        [Model Distillation, node-middle, edge=node,text width=9.3em
        ]
    ]
    [Generative Models, graph-middle, edge=graph, text width=6.8em
        [Embedding/Feature Generation, graph-middle, edge=graph, text width=9.3em
        ]
        [Generative Adversarial Network, graph-middle, edge=graph, text width=9.3em
        ]
        [Variational Autoencoder, graph-middle, edge=graph, text width=9.3em
        ]
        [Stable Diffusion, graph-middle, edge=graph, text width=9.3em
        ]
        [Model Inversion, graph-middle, edge=graph, text width=9.3em
        ]
    ]
    [Ensemble Methods, graph-middle1, edge=graph1, text width=6.8em
        [Static Ensemble, graph-middle1, edge=graph1, text width=9.3em
        ]
        [Adaptive Ensemble, graph-middle1, edge=graph1, text width=9.3em
        ]
    ]
  ]  
\end{forest}
\caption{A basic taxonomy of one-shot federated learning techniques. Note that Some hybrid methods employ multiple techniques.}
\label{fig:taxonomy-techniques}
\end{figure}

\section{One-shot Federated Learning Techniques}
\label{sec:method}
Figure~\ref{fig:taxonomy-techniques} illustrates our new taxonomy of current OFL techniques. These techniques aim to enhance the final global model's accuracy in light of the two challenges we previously mentioned. From an innovative perspective, we comprehensively categorize all the current methods into four main groups based on the techniques they employ: \textbf{Parameter Learning, Knowledge Distillation, Generative Models, and Ensemble Methods}. Besides these four techniques, many methods are hybrid methods, which employ a combination of multiple techniques.

\subsection{Parameter Learning}
Methods in the parameter learning category are derived from approaches like FedAvg and tend to directly learn the statistical information of local clients' model parameters to determine the global model's parameters.

\textbf{Clustering Method.} k-FED~\cite{dennis2021heterogeneity} employs Lloyd’s k-means clustering method to first cluster the information from local models and then uploads the cluster means to the server. This aids the server in learning more local information in a one-shot scenario, albeit with increased communication costs. \textbf{Regularizer.} MA-Echo~\cite{su2023one} considers aggregation of local models' parameters layer-wise and uses a unique method by introducing new norms. This approach helps the global model better account for differences among local models. \textbf{Conformal Prediction.} FedCP-QQ~\cite{humbert2023one} utilizes a conformal prediction method, demonstrating that for any distribution, a prediction set with the desired coverage can be computed in a single round of communication, thereby improving the global model's performance. \textbf{Bayesian Method.} FedFisher~\cite{jhunjhunwala2023towards} is an aggregation method that uses the empirical Fisher information matrix obtained by layer-wise Laplace approximation. This method first leverages Bayesian techniques to better model the posteriors of local clients and then aggregates for the global model. FedLPA~\cite{liufedlpa} employs a similar approach to FedFisher, further directly training the global model's parameters to achieve results. $\bm{\beta}$-PredBaye~\cite{hasan2024calibrated} first collects parameters from the posteriors of local clients and then introduces a tunable hyper-parameter $\bm{\beta}$. This parameter interpolates between a mixture and a product of the predictive posteriors by considering merging Gaussians in the predictive space, and aggregation is performed based on this approach. \textbf{Prototype Learning.} In prototype learning~\cite{snell2017prototypical}, one or more prototypes for each class are calculated using the samples from the training set. These prototypes can be considered as central representations summarizing and representing each class. PNFM~\cite{yurochkin2019bayesian} employs Bayesian Nonparametric Methods for neural matching based on prototype learning. FedMA~\cite{wang2020federated} extends the approach from PNFM by moving beyond fully-connected networks. It constructs a shared global model in a hierarchical manner by matching and averaging hidden elements (such as channels in convolutional layers or hidden states in LSTMs) with similar feature extraction characteristics, thus expanding the methodology to CNNs and LSTMs.

Since methods with parameter learning techniques are based on parameter optimization, they typically provide theoretical proofs of convergence. As these methods require uploading local model parameters, they can introduce privacy concerns. As a result, many approaches include ablation studies that integrate with Differential Privacy (DP). This category of methods can address the data heterogeneity challenge, potentially improving the global model's accuracy over traditional FL methods. However, due to the inherent limitations in accurately modeling the statistics, there is still a need to explore better approaches for capturing the features of local clients' parameters. Additionally, parameter learning techniques do not address the model heterogeneity challenge. Note that within parameter learning, prototype learning holds the most promise for integration with others.

\begin{table*}[!ht]
\centering
\resizebox{\textwidth}{!}{
\begin{tabular}{lcccc}
\toprule
\textbf{Method}  &\textbf{Distillation}    &\textbf{Generative Models} &\textbf{Ensemble} & \textbf{Additional Data}  \\
\midrule
DOSFL~\cite{zhou2020distilled} & Data & -& -&-\\
FedD3~\cite{song2023federated}& Data &- &- &-\\
\hdashline[0.5pt/5pt]
Fusion Learning~\cite{kasturi2020fusion}& -&Data Distribution Digest& -&-\\
XorMixFL~\cite{shin2020xor} & -&Embedding &- &Data Labels\\
FedOCD~\cite{liu2024fedocd} &- &Embedding & -&-\\
FedPFT~\cite{beitollahi2024parametric}& -&Feature Extractor (GMM) &- &Three Foundation Models\\
OSGAN~\cite{kasturi2023osgan} & -&Generative Adversarial Network&- &-\\
FedDEO~\cite{yang2024feddeo} &- &Stable Diffusion& -&-\\
FedBiP~\cite{chen2024fedbip} &- &Stable Diffusion& -&Latent Diffusion Model~\cite{rombach2022high}\\
FedDISC~\cite{yang2024exploring}& -&Stable Diffusion&- &CLIP~\cite{radford2021learning}\\
FedMITR~\cite{hao2025exploring}&- &Model Inversion	&-&-\\	
\hdashline[0.5pt/5pt]
DeDES~\cite{wang2023data} &- &- & Static&-\\
FedTMOS~\cite{qi2025fedtmos}&- & -&Static &-\\	
FENS~\cite{allouah2024revisiting} &- &- &Adaptive &-\\
FuseFL~\cite{tang2024fusefl} &- & -&Adaptive &-\\	
HPFL~\cite{shen2025bridging} &- & -&Adaptive &-\\	
Mediator~\cite{lai2025mediatormemoryefficientllmmerging} &- & -&Adaptive &-\\	
\hdashline[0.5pt/5pt]
FedOV~\cite{diao2023towards} & -&Generative Adversarial Network &Static&-\\
IntactOFL~\cite{zeng2024one} & -&Generative Adversarial Network &Adaptive&-\\
FedSD2C~\cite{zhang2024one} & Data& Variational Autoencoder& - &-\\
One-shot FL~\cite{guha2019one}	&Model $\rightarrow$&- &Static &Public Dataset\\
FedKT~\cite{li2020practical}	&Model $\rightarrow$ &- &Static &Public Dataset\\
FedKD~\cite{gong2022preserving}&Model $\leftarrow$ &- &Static &Public Dataset\\
Dense~\cite{zhang2022dense}&Model $\leftarrow$ &Generative Adversarial Network &Static &-\\
Co-Boosting~\cite{dai2024enhancing}&Model $\leftarrow$ &Generative Adversarial Network &Adaptive &-\\
FedCVAE~\cite{heinbaugh2023data} &Model&Variational Autoencoder&Static&Data Labels\\
\bottomrule
\end{tabular}
}
\caption{A synoptic overview of the surveyed solutions includes \textbf{Knowledge Distillation}, \textbf{Generative Models}, \textbf{Ensemble Methods}, and \textbf{Hybrid Methods}. The dashed line distinguishes between different categories. ``Additional data" refers to the extra data requirements needed. In the distillation column, ``$\rightarrow$" indicates that distillation is performed before applying the ensemble method, whereas ``$\leftarrow$" suggests the opposite, with distillation based on the ensemble model.}
\label{tab:hybrid}
\end{table*}

\subsection{Knowledge Distillation}
Knowledge distillation is a compression technique aimed at reducing the model's size and computational demands while maintaining as much accuracy as possible. It has been shown to address OFL problems effectively and is categorized into two major types: data distillation and model distillation.

\textbf{Data Distillation.} DOSFL~\cite{zhou2020distilled} involves distilling the local private dataset at each client and then training on the distilled data at the server. FedD3~\cite{song2023federated} operates in resource-constrained edge environments by collecting decentralized dataset distillation to train the global model. Additionally, FedSD2C~\cite{zhang2024one} incorporates data distillation alongside other techniques, which we will analyze in Section~\ref{sec:hybrid}. \textbf{Model Distillation.} The methods utilize model distillation, and as shown in Table~\ref{tab:hybrid}, these often incorporate multiple techniques, typically in combination with ensemble methods. Generally, there are two approaches: (1) conducting model distillation first and then ensembling~\cite{guha2019one,li2020practical}, (2) or vice versa—ensembling first and distilling the ensembled model afterward~\cite{gong2022preserving,zhang2022dense,dai2024enhancing}. We will discuss these methods in detail in Section~\ref{sec:hybrid}.

Compared to parameter learning techniques, knowledge distillation provides better privacy preservation. By allocating more computational resources to distillation, it iteratively enhances the global model's performance, yielding higher accuracy and supporting model heterogeneity.

\subsection{Generative Models}
Generative models are a widely used class of techniques in machine learning, designed to train synthetic samples that closely resemble the original data, thereby ensuring similar data distributions. Consequently, generative models can be utilized in OFL to aid the server in training on the synthetic datasets, which is an effective learning from local datasets. \textbf{Embedding/Feature Generation.} In Fusion Learning~\cite{kasturi2020fusion}, the distribution digest of client data and their local model parameters are sent to the server. The server regenerates dataset based on these distribution digest, integrates parameters from multiple devices to construct a global model, trains the model using the combined dataset. XorMixFL~\cite{shin2020xor} assumes both the global server and clients possess labeled samples from a global class. The server collects encoded data samples (embeddings) from other devices to construct the global data samples. It employs the exclusive OR operation (XOR) to protect privacy during the encoding process. FedOCD~\cite{liu2024fedocd} focuses on cross-domain recommendation and applies local differential privacy for security while processing the generated user embeddings. FedPFT~\cite{beitollahi2024parametric} utilizes a feature extractor to derive the corresponding Gaussian mixture models (GMMs) to model synthetic sample features, transferring per-client parametric models.
\textbf{Generative Adversarial Networks.} OSGAN \cite{kasturi2023osgan} further employs Generative Adversarial Networks (GANs) as a generative model to enhance performance. Other uses~\cite{zhang2022dense,diao2023towards,zeng2024one,dai2024enhancing} of Generative Adversarial Networks, along with additional techniques, can be found in Table~\ref{tab:hybrid} and will be discussed in detail in Section~\ref{sec:hybrid}.
\textbf{Variational Autoencoder.} Researchers have further utilized Variational Autoencoders in approaches such as FedCAVE~\cite{heinbaugh2023data} and FedSD2C~\cite{zhang2024one}.
\textbf{Stable Diffusion.} FedDEO~\cite{yang2024feddeo} initially adopts popular diffusion models, transmitting image captions (local descriptions) to better capture the distribution of client local data. This assists the server in generating synthetic data that align more closely with the client's data distribution. FedBiP~\cite{chen2024fedbip} further utilizes a better Stable Diffusion model, specifically the foundation model (Latent Diffusion Model~\cite{rombach2022high}), to improve performance. FedDISC~\cite{yang2024exploring} leverages a pre-trained model CLIP~\cite{radford2021learning} with prototype learning. It involves four steps: Prototype Extraction, Pseudo Labeling, Feature Processing, and Image Generation, which lead to improved outcomes. 
\textbf{Model Inversion.} FedMITR~\cite{hao2025exploring} employs model inversion techniques to obtain statistical information about some private data within local models. Additionally, it uses a vision transformer to perform token relabeling.

Based on the above discussion, we find that compared to parameter learning, knowledge distillation and generative models offer advantages beyond higher accuracy when given more computational resources. We observe that both techniques address data and model heterogeneity challenges. Additionally, they enhance security by not directly sharing client parameters. Since both techniques aim to statistically approximate the true local datasets without exposing actual data, as local models also reflect local datasets. Thus, they enhance the global model's capability with similar approaches to learning client-specific features, which enables the global model to train more effectively while mitigating out-of-distribution (OOD) issues.

These two techniques also have their differences. Model distillation improves methods by distilling parameter information from clients rather than using their raw data. In contrast, generative models enhance learning by generating synthetic data that capture the characteristics of clients' private datasets. Although data distillation also emphasizes the data aspect, generative models can address the inefficiencies associated with knowledge distillation methods, often leading to superior performance. As a result, numerous researchers are actively pursuing advancements in methods that better reflect local data distributions, especially through generative models. This progression is evident as the focus has shifted from GANs to VAEs, then to Stable Diffusion, and finally to increasingly sophisticated diffusion models.

\subsection{Ensemble Methods}
\label{sec:ensemble}
Ensemble methods represent one of the most straightforward solutions to addressing OFL challenges (data heterogeneity and model heterogeneity). The most na\"ive ensemble approach involves aggregating global models obtained from local clients. However, ensemble models can actually be combined with a wide range of other techniques. Currently, these combinations are mainly categorized into two types: static and adaptive ensemble methods.

\textbf{Static.} Static ensemble methods are typically used to perform operations like averaging (DeDES~\cite{wang2023data}) or taking the maximum based on the outputs from local clients. FedTMOS~\cite{qi2025fedtmos} enhances this approach by integrating a lightweight reinforcement learning model, Tsetlin Machine, with prototype learning. Each client can train a different Tsetlin Machine for each class. Due to the unique nature of the Tsetlin Machine, which employs Boolean feature representation, it is well-suited for handling sparse or high-dimensional data. This leads to improved results on the server side. \textbf{Adaptive.} FENS~\cite{allouah2024revisiting} balances communication rounds and accuracy by adjusting the ensemble weights based on the similarity between global and local models. From the causality perspective, FuseFL~\cite{tang2024fusefl} identifies that the performance drop comes from the isolated training problem. Then, FuseFL proposes a bottom-up approach that trains and merges sub-models adaptively to enhance invariant feature learning, thus alleviating spurious fitting. HPFL~\cite{shen2025bridging} proposes regarding the personalized modules as hot pluggable plug-ins which can be selected according to the input data during inference. The plug-ins only need to be sent to server once. Mediator~\cite{lai2025mediatormemoryefficientllmmerging} aggregates different finetuned LLMs that can be trained by different parties with low memory occupation and adaptive routing and compression.

Other methods~\cite{guha2019one,li2020practical,gong2022preserving,zhang2022dense,heinbaugh2023data,diao2023towards,zeng2024one,dai2024enhancing} that utilize ensemble techniques are listed in Table~\ref{tab:hybrid} and will be discussed in Section~\ref{sec:hybrid}. While ensemble methods directly leverage local models, they can be combined with other techniques to ensure privacy and improve accuracy. We also find that adaptive ensemble methods typically yield better results.

\subsection{Hybrid Methods}
\label{sec:hybrid}

As previously mentioned, current papers do not adequately classify methods addressing one-shot Federated Learning. Table~\ref{tab:hybrid} illustrates the challenges of classification and highlights the effectiveness of our unique classification approach. This paper thoroughly examines all current methods of OFL from a technical contribution perspective. A single method can employ multiple techniques. The three techniques that can tackle model heterogeneity—knowledge distillation, generative models, and ensemble methods, are often integrated to enhance performance and privacy. These methods can be combined in pairs or even all three together. Based on this, and following our detailed discussions of each technique, we summarize these methods, providing an in-depth analysis of how their integration enhances performance.

\textbf{Integrating Generative Models with Ensemble Methods.} FedOV~\cite{diao2023towards} addresses the open-set problem with an OOD perspective, where some classes may be missing on certain clients due to label skew. To handle this, it uses GAN to generate outliers with feature corruption and employs adversarial training techniques. The generated data samples serve as additional unknown classes during local training, and the results are obtained by ensemble methods from local clients. IntactOFL \cite{zeng2024one} also employs GANs to generate synthetic data samples but adopts a more flexible approach by utilizing a mixture of experts (MoE) model~\cite{jacobs1991adaptive}. This allows for dynamic adjustment of the ensemble weights for each local client, tailored specifically to each data instance, which greatly improves the performance.

\textbf{Integrating Generative Models with Distillation.} FedSD2C~\cite{zhang2024one} begins by using a pretrained Autoencoder to extract distillate synthesis. To reduce communication costs and enhance the privacy of the distillates, clients perturb the distillates using the Fourier transform. To ensure that the distillates fully encompass local information, clients perform a V-information based Core-Set selection when utilizing the encoder. Finally, the server collects all the distillates and uses the pretrained autoencoder's decoder to train an accurate global model.

\textbf{Integrating Ensemble Methods with Distillation.} One-shot FL~\cite{guha2019one}, particularly in the context of semi-supervised learning, leverages distillation to reduce the size of the global model and enhance privacy guarantees. The approach then applies average ensemble methods to distilled local models to achieve favorable results. FedKT~\cite{li2020practical}, while using a similar approach, extends previous methods from support vector machines (SVMs). It incorporates a hierarchical knowledge transfer framework with a two-tier privacy aggregation of teacher ensembles (PATEs) structure to improve accuracy and introduces the concept of consistent voting to strengthen the ensemble. Both of these methods utilize publicly available datasets to further enhance precision. FedKD~\cite{gong2022preserving} makes use of unlabeled, cross-domain public data and requires transferring products of these. To address privacy concerns, it incorporates a quantized and noisy ensemble into local models. Following the ensemble process, it applies distillation to derive the final global model. Notably, by employing a combination of both shared techniques and additional public datasets, the results demonstrate that first ensembling and then distilling yields better outcomes. This is likely because, as mentioned, distillation can be inherently inefficient and may result in some loss of information. Distilling before ensembling might lead to models that are not as accurate prior to aggregation.

\textbf{Combination with Three Techniques.} Dense~\cite{zhang2022dense} uses an ensemble of local models uploaded by clients to train a GAN-type generator, which then generates synthetic data samples. The knowledge from the ensemble models is distilled into the global model, which is also trained using the synthetic data to enhance the accuracy. Co-Boosting \cite{dai2024enhancing} employs a similar generator and also follows the process of ensembling before distillation. However, it extends Dense by introducing a reinforcing approach. This method continuously adjusts the weights of the ensemble model's local models based on synthetic data, and through this process, it finetunes over multiple iterations. Similar to FuseFL~\cite{tang2024fusefl}, although model parameters are uploaded only once, the method undergoes several iterations of optimization, achieving high accuracy. Unlike the previous two methods, in FedCVAE~\cite{heinbaugh2023data}, each local client trains an Autoencoder and transmits the local label information and decoder. The server then uses the uploaded client decoder parameters and local label distributions to train a server decoder. Ultimately, synthetic samples from the server decoders are used to train the global model. FedCAVE-KD employs knowledge distillation of local decoders to train the server decoder, while FedCAVE-Ens ensembles client decoders and trains the server decoder.

To avoid repetition in Section~\ref{sec:ensemble}, we address the privacy issues associated with ensemble methods here, as they often combine with other techniques. For example, using distillation before ensembling can effectively enhance privacy. Conversely, ensembling first and then distilling can achieve security by adding noise to the uploaded local models, similar to parameter learning. Therefore, regardless of how these techniques are combined, we find they all can ensure privacy.

\section{Discussions and Future Directions}
\label{sec:future}
In this section, having discussed all the current techniques, we note that these methods have successfully implemented one-shot learning, reduced communication overhead, and ensured data privacy. We will summarize the findings based on these methods from the perspective of improving accuracy. Following this, we will present a blueprint for future directions that focuses not only on test accuracy but also considers other aspects, grounded in practical guidance.

\subsection{Findings}

~~~~\textbf{Prototype learning has a bright future.} Parameter learning faces challenges in addressing model heterogeneity and is generally limited by the method's capacity to fully capture the weights in local parameters and discard spurious correlations caused by data heterogeneity, which can restrict accuracy. Consequently, compared to other methods, parameter learning does not demonstrate significant advantages. However, fine-grained approaches such as prototype learning within parameter learning show substantial potential. They enable feature extraction by class, thereby enriching the global model's information. These approaches can be readily integrated with other methods, as seen with FedTMOS, which employs an ensemble method, and FedDISC, which utilizes generative model diffusion. Thus, employing prototype learning or conducting more granular analyses of local models or class features in OFL holds significant promise.

\textbf{Better Generative Model is needed.} Generative models tend to perform better compared to knowledge distillation methods, primarily because they typically operate directly on the data rather than the model. Even when considering data distillation as opposed to model distillation, it becomes challenging to compare with generative methods due to the compression involved in distillation, which limits accuracy. We also observe that the academic community is striving for improved generative models. Therefore, if better generative models become available, they should be utilized in the OFL method. Of course, it is also worth experimenting with improved distillation methods for one-shot Federated Learning.

\textbf{Adaptive ensemble method shows great potential.} Compared to static ensembles, adaptive ensembles achieve better results. Both FENSE~\cite{allouah2024revisiting} and Co-boosting~\cite{dai2024enhancing} consider the differences between global and local models, allowing for a refinement of ensemble weights for local models to enhance accuracy. However, these dynamic methods still struggle to adjust according to specific data samples. Due to data heterogeneity, the disparity between data samples across different clients can be significant, necessitating varied weight distributions for different samples. IntactOFL~\cite{zeng2024one} significantly improves accuracy using an MoE approach, although it uses a basic generative model like GAN and does not use distillation. Its accuracy surpasses both Dense and Co-boosting. Consequently, a better adaptive ensemble method in OFL that accounts for data samples will likely yield better results.

\textbf{Uploading once but updating multiple iterations.} We found that although OFL restricts the upload to a single round, performing multiple local updates can better leverage the upload. Aside from Co-boosting~\cite{dai2024enhancing}, FuseFL~\cite{tang2024fusefl} updates one block (one or multiple layers) per iteration from the bottom up view, thereby updating from a causal perspective, ultimately achieving excellent results. These findings suggest that a multi-shot approach should be favored over a one-shot approach on the server while keeping the same communication costs.

\textbf{Aggregating more techniques is preferable.} The results from the papers we previously discussed demonstrate that, beyond employing a better model within various techniques, there is no single ``silver bullet" for improving one-shot Federated Learning outcomes. Instead, integrating multiple techniques can lead to better results. In detail, incorporating prototype learning when training local models can significantly refine their representation capabilities. Utilizing more fine-grained knowledge distillation when distilling adaptive ensemble models can further enhance the precision and effectiveness of the ensembles. Additionally, training the distilled global model with synthetic datasets generated by superior generative models can improve its robustness and generalization, thereby contributing to enhanced overall performance. Besides these approaches, an ideal method should involve conducting multiple iterations of local updates. In this scenario, the uploaded statistical information can be maximized, leading to the best possible results.

\subsection{Future Directions}
In addition to the findings we have uncovered that can guide enhancements in the accuracy of OFL methods, this subsection will provide a practical roadmap for the future development of one-shot Federated Learning.

\textbf{Data-free Requirement.} XorMixFL~\cite{shin2020xor} and FedCAVE~\cite{heinbaugh2023data} both require the transmission of local clients' data labels, posing a risk of data privacy leakage. On the other hand, methods like One-shot FL~\cite{guha2019one}, FedKT~\cite{li2020practical}, and FedKD~\cite{gong2022preserving} depend on additional public datasets; similarly, approaches such as FedPFT~\cite{beitollahi2024parametric}, FedBiP~\cite{chen2024fedbip}, and FedDISC~\cite{yang2024exploring} utilize foundation models, leveraging extra public datasets as well. However, OFL ideally should be data-free; using additional datasets might lead the global model to learn biased information that does not align with the target dataset, such as certain industrial or biological datasets, which may ultimately reduce accuracy. Therefore, data-free operation should be one of the key objectives for OFL methods.

\textbf{Scalability for LLMs.} Training LLMs in FL with multiple rounds requires significant communication costs because of the enormous model sizes~\cite{openfedllm,tang2024fusionllmdecentralizedllmtraining}. OFL has significant potential to reduce this communication cost, thus implementing training or finetuning LLMs across different geo-distributed clients. However, most current experimental approaches for OFL methods primarily focus on models like LeNet, VGGNet, and ResNet, with datasets mostly comprising MNIST, CIFAR, Tiny ImageNet, or simple medical datasets and etc. There is some work trying to merge different finetuned LLMs~\cite{lai2025mediatormemoryefficientllmmerging} only once. Besides, state-of-the-art OFL methods can accommodate model heterogeneity~\cite{li2021fedbn,shen2025bridging}, knowledge distillation between heterogeneous LLMs~\cite{gu2024minillm} and collaboration between LLMs~\cite{wang2024mixture} can be implemented in OFL. In the future, the FL community could explore OFL methods to enable more practical training of LLMs with different parties.

\textbf{Practical Applications.} Based on the experimental considerations mentioned above, besides the LLM field, these methods should be applicable to more practical applications. For instance, combining them with vertical federated learning (VFL) could result in one-shot VFL methods. The OFL approach can be seen in applications like FedISCA~\cite{kang2023one} with biomedical data, FedD3~\cite{song2023federated} focusing on resource-constrained edge environments, and OFL-W3~\cite{jiang2024ofl} integrated with blockchain and Web 3.0 technology. Given the one-shot nature of OFL, which eliminates the need for synchronization, it can also be widely utilized in cloud-edge collaborative frameworks. Additionally, there is potential for application in emerging fields like sequential federated learning~\cite{wang2024one}. These highlight the broad application prospects of OFL.

\textbf{Advanced Optimizations.}  Although OFL significantly reduces communication overhead, integrating it with LLMs and cloud-edge collaborative frameworks presents additional challenges. These challenges arise because such models are large and often support numerous services, leading to substantial parameter sizes even in a single round of interaction. Therefore, further optimization is required, such as implementing model compression~\cite{hu2024practical} or acceleration techniques~\cite{liu2024accelerate}.

In summary, while OFL methods demonstrate tremendous practical potential, improvements should not be limited to enhancing accuracy. Consideration should also be given to the aspects we mentioned. By addressing these considerations and integrating our findings, we outline a clear direction for the future development of OFL methods.

\section{Conclusion}
\label{sec:conclusion}
In this paper, we focus on the novel distributed machine learning paradigm, one-shot Federated Learning. We provide a detailed analysis of the challenges faced within one-shot Federated Learning, propose an innovative taxonomy, and thoroughly discuss the specifics of various methods based on this classification. Our paper comprehensively covers all existing OFL literature, comparing the advantages, developmental trajectories, and areas for improvement across different techniques. Based on this analysis, we discuss our findings and offer numerous future directions from a practical standpoint for this rapidly evolving field. This survey aims to consolidate existing knowledge and lay a foundation for future work in the promising one-shot Federated Learning area.

\section*{Acknowledgements}
Dr. Jialin Li is supported by the Singapore Ministry of Education, Academic Research Fund Tier 1 (T1 251RES2104) and Tier 2 (MOE-T2EP20222-0016). 

\bibliographystyle{named}

\end{document}